\documentclass[letterpaper, 10 pt, conference]{ieeeconf}
\IEEEoverridecommandlockouts
\usepackage{cite}
\usepackage{amsmath,amssymb,amsfonts}
\usepackage{algorithmic}
\usepackage{graphicx}
\usepackage{textcomp}
\usepackage{array}
\usepackage{booktabs}
\usepackage{rotating}
\usepackage{xcolor}
\def\BibTeX{{\rm B\kern-.05em{\sc i\kern-.025em b}\kern-.08em
    T\kern-.1667em\lower.7ex\hbox{E}\kern-.125emX}}

\begin{document}




\title{\LARGE \bf Panoster: End-to-end Panoptic\\
Segmentation of LiDAR Point Clouds}

\author{Stefano Gasperini$^{1,2}$ \quad Mohammad-Ali Nikouei Mahani$^{2}$ \quad Alvaro Marcos-Ramiro$^{2}$\\Nassir Navab$^{1,3}$ \quad Federico Tombari$^{1,4}$
\thanks{$^{1}$ Computer Aided Medical Procedures, Technical University of Munich, Munich, Germany. \textit{stefano.gasperini@tum.de}}
\thanks{$^{2}$ BMW Group, Munich, Germany.}
\thanks{$^{3}$ Computer Aided Medical Procedures, Johns Hopkins University, Baltimore, Maryland, USA.}
\thanks{$^{4}$ Google, Zurich, Switzerland.}
\thanks{This is a preprint of the article accepted at Robotics and Automation Letters (RA-L). \textcopyright \thinspace 2021 IEEE.}
}

\maketitle


\label{parts:abstract}

\begin{abstract}

Panoptic segmentation has recently unified semantic and instance segmentation, previously addressed separately, thus taking a step further towards creating more comprehensive and efficient perception systems.
In this paper, we present Panoster, a novel proposal-free panoptic segmentation method for LiDAR point clouds.
Unlike previous approaches relying on several steps to group pixels or points into objects, Panoster proposes a simplified framework incorporating a learning-based clustering solution to identify instances.
At inference time, this acts as a class-agnostic segmentation, allowing Panoster to be fast, while outperforming prior methods in terms of accuracy.
Without any post-processing, Panoster reached state-of-the-art results among published approaches on the challenging SemanticKITTI benchmark, and further increased its lead by exploiting heuristic techniques.
Additionally, we showcase how our method can be flexibly and effectively applied on diverse existing semantic architectures to deliver panoptic predictions.

\end{abstract}



\section{Introduction}
\label{sec:intro}


Scene understanding is a fundamental task for autonomous vehicles.
Panoptic segmentation (PS)~\cite{kirillov2019panoptic} combines semantic and instance segmentation, enabling a single system to develop a more complete interpretation of its surroundings.
PS distinguishes between \textit{stuff} and \textit{thing} classes~\cite{kirillov2019panoptic}, with the former being uncountable and amorphous regions, such as \textit{road} and \textit{vegetation}, and the latter standing for countable objects, such as \textit{people} and \textit{cars}.
The two have been addressed separately for decades, leading to significantly different approaches: \textit{stuff} classifiers are typically based on fully convolutional networks~\cite{long2015fully}, while \textit{thing} detectors often rely on region proposals and the regression of bounding-boxes~\cite{ren2015faster}.
Tackling semantic and instance segmentation jointly with PS
allows to save costly and limited resources~\cite{mohan2020efficientps}.

Since the introduction of PS,
several approaches have been proposed to address this new task. While most focused on images~\cite{porzi2019seamless, mohan2020efficientps, cheng2020panoptic} or RGBD data~\cite{wang2019associatively, pham2019jsis3d}, only a handful of methods used LiDAR point clouds~\cite{hurtado2020mopt, milioto2020lidar}, none of which has managed to outperform systems combining separate networks for semantic segmentation and object detection~\cite{behley2020benchmark}.
LiDAR sensors have proven to be particularly useful for self-driving cars, capturing precise distance measurements of the surrounding environment.
Segmenting LiDAR point clouds is a crucial step for interpreting the scene.
Compared to image pixels, LiDAR point clouds pose various challenges: they are unstructured, unordered, sparse, and irregularly sampled.

\begin{figure}[t]
\centering
  \includegraphics[width=0.48\textwidth]{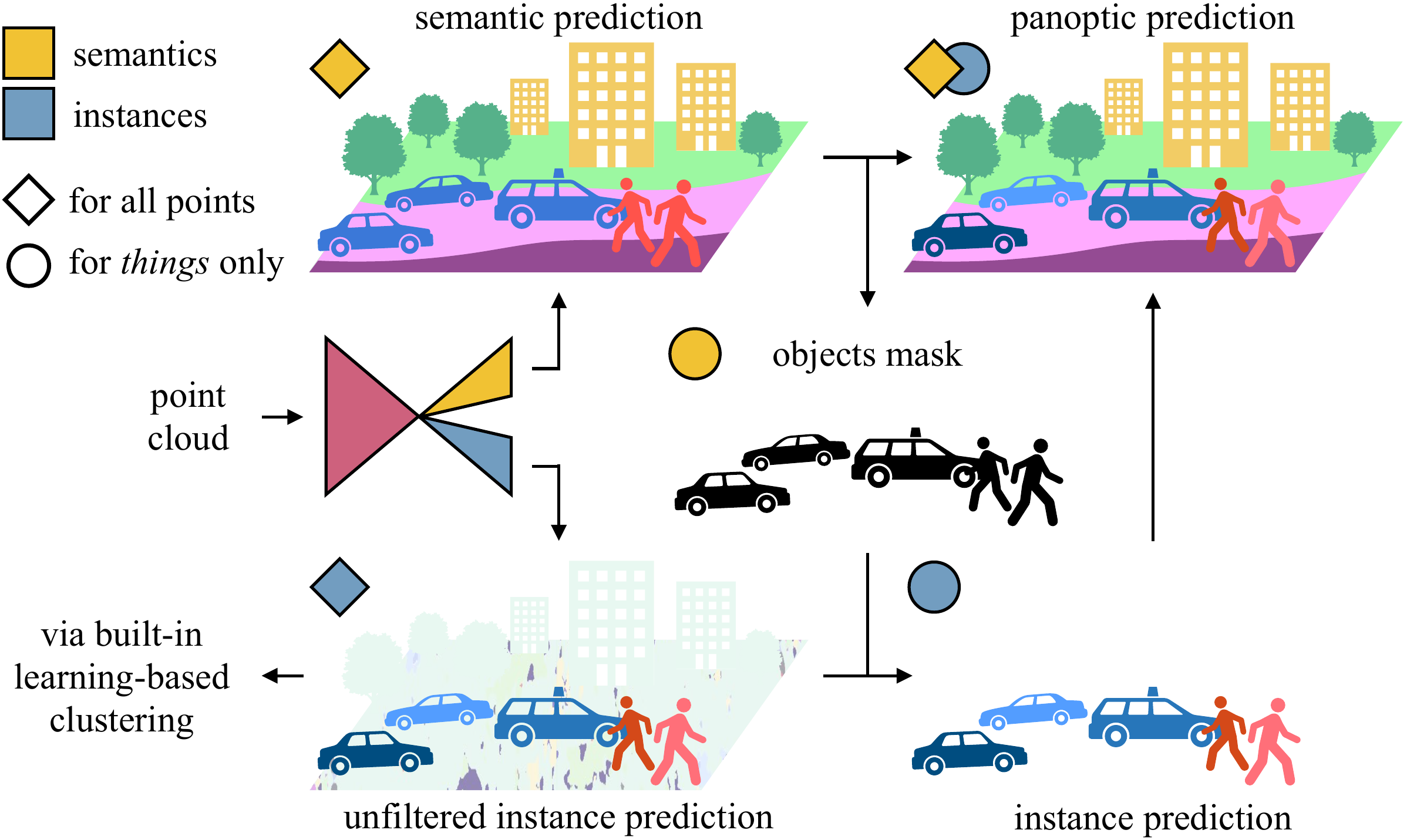}
   \caption{Proposed Panoster framework. A semantic class and an instance ID are predicted for every point, then filtered and combined for panoptic segmentation.}
   \label{fig:framework}
\end{figure}

In this paper, we introduce a novel proposal-free approach for panoptic segmentation of LiDAR point clouds. We name our method Panoster, standing for panoptic clustering.
Panoster brings panoptic capabilities
to semantic segmentation networks
by adding an instance branch trained to resolve objects with a learning-based clustering method adapted from~\cite{gasperini2020signal}.
Unlike existing proposal-free approaches~\cite{cheng2020panoptic, milioto2020lidar}, Panoster, thanks to its clustering solution integrated in the model, does not need further groupings to form objects, as it outputs instance IDs straight away from the network.
The contributions of this paper can be summarized as follows:
\begin{itemize}
    \item We introduce a new proposal-free approach for panoptic segmentation, fully end-to-end, flexible and fast, with a small overhead on top of an equivalent semantic segmentation architecture.
    \item To the best of our knowledge, for the first time in panoptic segmentation, we exploit an entirely learning-based end-to-end clustering approach to separate the instances, resulting in a fully end-to-end panoptic method.
    \item Our novel instance segmentation acts as class-agnostic segmentation, has no thresholds, and delivers instance IDs directly, without requiring further grouping steps.
\end{itemize}    
We apply Panoster on the challenging task of panoptic segmentation for LiDAR 3D point clouds, outperforming in a fraction of the time state-of-the-art methods.
Furthermore, we show the flexibility of Panoster by deploying it on both projection-based and point-based architectures, making it the first panoptic method to consume LiDAR 3D points directly.


\section{Related Work}
\label{sec:related}


In this section, we provide a brief overview of existing approaches,
and highlight overlaps and differences with Panoster.

\subsection{Semantic Segmentation}
Depending on the input representation, semantic segmentation methods can be grouped into projection-based and point-based.
%
Projection-based approaches eliminate the challenging irregularities of 3D points by projecting them into an intermediate grid. This grid representation can be made of voxels~\cite{maturana2015voxnet, riegler2017octnet} or pixels~\cite{milioto2019rangenet++, cortinhal2020salsanext}, allowing the use of 3D or 2D convolutions respectively.
While the former projection tends to be highly inefficient for this task, the latter can count on state-of-the-art methods from image segmentation, albeit discarding valuable geometric information by transitioning from 3D to 2D.

PointNet~\cite{qi2017pointnet} pioneered point-based methods and deep learning on point clouds. They used multilayer perceptrons (MLP) to learn features from each point, then aggregated with a global max-pooling. Additional MLP-based architectures were proposed in~\cite{qi2017pointnet++, hu2020randla, lang2019pointpillars}.
More recent approaches, such as KPConv~\cite{thomas2019kpconv} and SpiderCNN~\cite{xu2018spidercnn} have focused on creating new convolution operations for points. 
KPConv~\cite{thomas2019kpconv} extended the idea of 2D image kernels to 3D points, by defining an operation which derives new 3D points from the multiplication of input 3D point features with flexible kernel weights.
Recently, hybrid approaches have been developed~\cite{kochanov2020kprnet, tang2020searching}, combining point and projection-based solutions.

Our method Panoster extends semantic segmentation approaches to panoptic segmentation. 
In this paper we apply Panoster on KPConv~\cite{thomas2019kpconv} and SalsaNext~\cite{cortinhal2020salsanext}, representative of point-based and projection-based categories respectively.

\subsection{Instance Segmentation}
Instance segmentation methods can be divided into detection-based and clustering-based. The former was established in the image domain by Mask R-CNN~\cite{he2017mask} which has two stages: region proposals are extracted, refined into bounding boxes, and segmented with a pixel-level mask to identify the objects. Mask R-CNN has catalyzed a variety of methods, some of which address panoptic segmentation~\cite{mohan2020efficientps, xiong2019upsnet, hurtado2020mopt}, as described in Section~\ref{sec:related_work_pano_seg}.
Clustering-based approaches learn an embedding space in which instances can be easily clustered~\cite{de2017inst_clust, neven2019instance}. Therefore, the segmentation is typically done by deploying a data analysis clustering technique, such as DBSCAN~\cite{ester1996dbscan}, on the latent features, which were learned to aid the separation.
These methods were extended to point clouds in~\cite{hou20193dsis, yi2019gspn, zhang2020instance}.
%

Panoster operates differently, by clustering object instances thanks to loss functions optimizing the network outputs directly. Although our instance segmentation is learning-based, it does not explicitly rely on the optimization of a latent space via embeddings, differentiating itself from clustering-based methods described above. Furthermore, Panoster performs panoptic segmentation, resulting in a more comprehensive approach.

\subsection{Panoptic Segmentation}
\label{sec:related_work_pano_seg}
PS methods can be divided into two categories: proposal-based and proposal-free.
The vast majority are from the image domain, of which the larger stake are proposal-based or top-down methods (e.g. UPSNet~\cite{xiong2019upsnet}, AdaptIS~\cite{sofiiuk2019adaptis}, Seamless~\cite{porzi2019seamless}, EfficientPS~\cite{mohan2020efficientps}): two-stage approaches, which are usually based on the well-known Mask R-CNN~\cite{he2017mask}.
They rely on the detection of \textit{things} first, followed by a refinement step and a semantic segmentation of \textit{stuff}. Their main drawbacks are the needs of resolving mask overlaps and conflicts between \textit{thing} and \textit{stuff} predictions.
Instead, proposal-free or bottom-up methods are single-stage and follow the opposite direction. They segment the scene semantically, and cluster the instances within the predicted \textit{thing} segments. DeeperLab~\cite{yang2019deeperlab} was the first of this kind, regressing objects center and corners from images. SSAP~\cite{gao2019ssap} used a cascaded graph partition module to group pixels into instances according to a pixel-pair affinity pyramid.
Panoptic-DeepLab~\cite{cheng2020panoptic} simplified this concept by predicting the instance center locations, and using a Hough-Voting scheme to group the pixels to the closest center.

While PS received a lot of attention in the image domain, PS on LiDAR point clouds is yet broadly unexplored, due to the only recent introduction of a suitable public dataset, i.e. SemanticKITTI with panoptic labels~\cite{behley2020benchmark}.
In~\cite{behley2020benchmark} strong baselines combined state-of-the-art semantic segmentation methods KPConv~\cite{thomas2019kpconv} or RangeNet++~\cite{milioto2019rangenet++}, with the object detector PointPillars~\cite{lang2019pointpillars}.
PanopticTrackNet~\cite{hurtado2020mopt} followed the approach of EfficientPS~\cite{mohan2020efficientps} and extended it
to consume LiDAR point clouds: it is top-down, based on Mask R-CNN~\cite{he2017mask} to resolve the instances. Additionally, a new tracking head delivers temporally consistent instance IDs.
The only existing bottom-up method for LiDAR point clouds is~\cite{milioto2020lidar}, which we denote as LPSAD. It is based on semantic embeddings and, as in Panoptic-DeepLab~\cite{cheng2020panoptic}, the regression of offsets to the object centers, then instances are extracted by an iterative clustering procedure, grouping points according to their center offset prediction. To date, no method managed to outperform the combined baselines proposed in~\cite{behley2020benchmark}.

Panoster is significantly different from existing panoptic approaches.
It is the first to use an entirely learning-based clustering technique to resolve the instances. As LPSAD~\cite{milioto2020lidar}, our method is bottom-up and applied on LiDAR point clouds, but compared to it, our instance branch outputs instance IDs directly, without requiring any subsequent grouping steps, rendering it fully end-to-end trainable.


\section{Panoster}
\label{sec:method}



\subsection{Overview}
\label{sec:overview}

\begin{figure}[t]
\centering
  \includegraphics[width=0.48\textwidth]{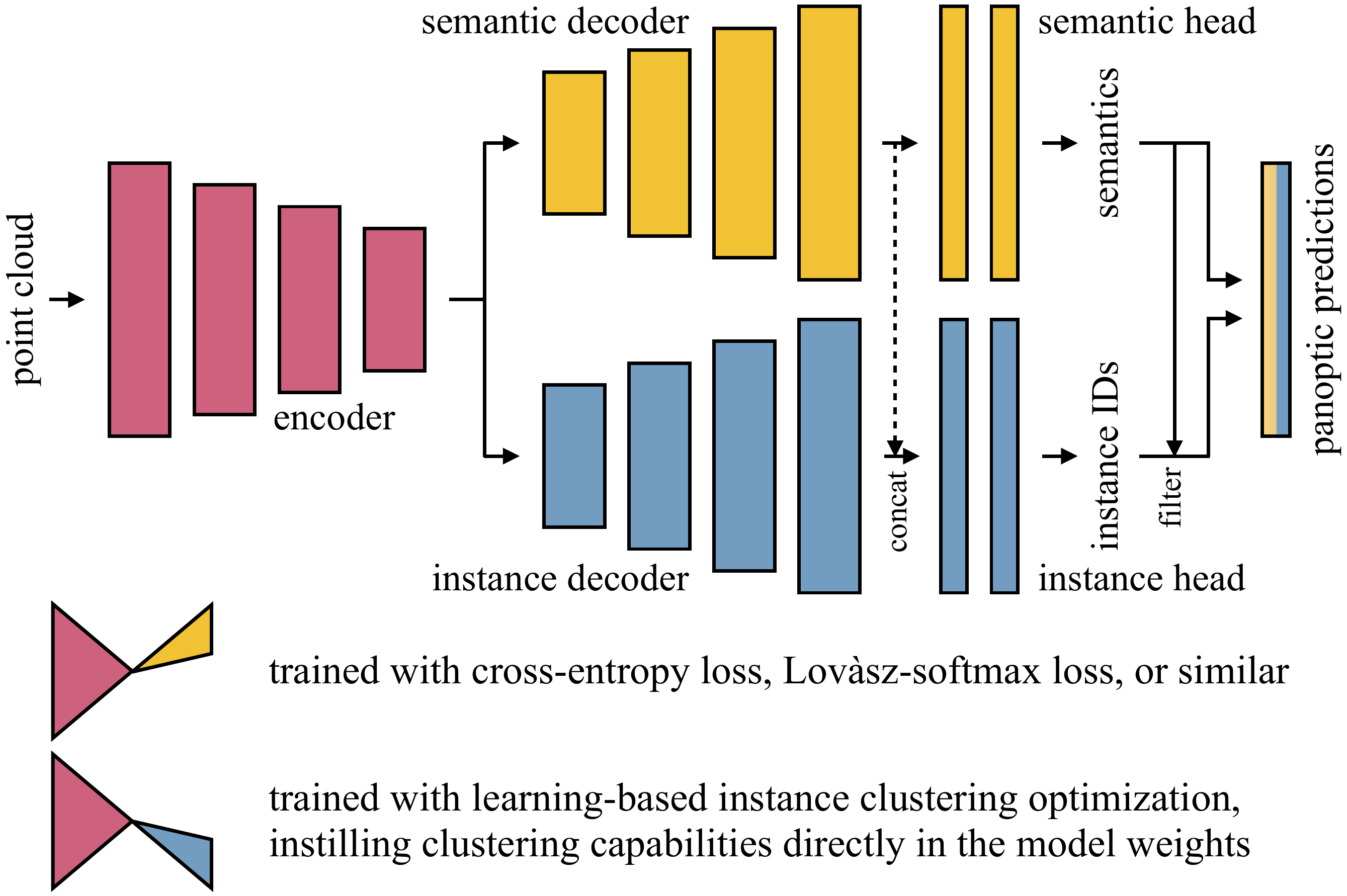}
   \caption{Panoster simplified architecture. The input is raw 3D points or a 2D spherical projection. During training, gradients do not back-propagate through the dashed line. At inference time, both decoders and heads process data in the same way.}
   \label{fig:architecture}
\end{figure}

As described in Section~\ref{sec:related_work_pano_seg}, current proposal-free panoptic approaches~\cite{cheng2020panoptic, milioto2020lidar} pair a network with an external grouping technique to form objects.
The latter exploits the network predictions, and is required in order to output any meaningful instance IDs.
With the proposed Panoster, we simplify by removing this extra step, and incorporating the clustering in the network itself.
We achieve this with an entirely learning-based approach, adapted from the signal processing domain~\cite{gasperini2020signal}, to output instance IDs directly, for each point given in input (Section~\ref{sec:inst_seg}).
To obtain panoptic predictions, it is then sufficient to filter out these IDs for all \textit{stuff} points (Section~\ref{sec:pano_seg}), as shown in Fig.~\ref{fig:framework}.
Thanks to custom loss functions, our integrated clustering method instills grouping capabilities right in the model weights.
At inference time, these layers process data in the same way as their semantic segmentation counterparts, while pursuing the rather different task of instance segmentation.
In Fig.~\ref{fig:architecture} we showcase Panoster general architecture, highlighting how each part is trained.

\subsection{Architecture}
\label{sec:architecture}

Panoster consists of the following modules: a shared encoder, decoupled symmetric decoders and heads for the two tasks, mask-based fusion and optional post-processing.

Motivated by~\cite{cheng2020panoptic, hurtado2020mopt, milioto2020lidar}, we opt for a shared encoder with two separate decoders, each developing its task starting from common features.
Following Fig.~\ref{fig:architecture}, the backbone, the semantic decoder and head follow the design of a standard semantic segmentation network, consuming a full 360 degree LiDAR point cloud.
In this paper we apply Panoster on KPConv~\cite{thomas2019kpconv} and SalsaNext~\cite{cortinhal2020salsanext}.
We refer to the two variants as PanosterK and PanosterS, respectively. 
Therefore, depending on the configuration, our framework can be fed with raw 3D points or with a 2D spherical projection.

As shown in Fig.~\ref{fig:architecture}, our method has a second decoder and head, dedicated to instance segmentation.
From the architecture perspective, the two branches are identical, except for the size of the last layer: the semantic branch output size depends on the number of semantic classes, while the output of the instance branch is sized $N$, which is the number of predictable instances (Section~\ref{sec:inst_seg}).
Additionally, the input of the instance head is the concatenation of both decoders outputs.
Apart from the input and output sizes of the heads, the two branches differ as each is optimized with its own loss function, which is designed to tackle its task.

We train the semantic branch to predict both \textit{stuff} and \textit{thing} classes. We use the same objective functions as in~\cite{cortinhal2020salsanext}, namely weighted cross-entropy, and Lov\`{a}sz-softmax loss~\cite{berman2018lovasz}, optimizing the Jaccard index.
We weight their sum 0.7 for PanosterK and 1.0 for PanosterS. Moreover, following~\cite{cheng2020panoptic}, to improve the accuracy on small or distant \textit{thing} objects, we triple the cross-entropy weight on instances smaller than 100 points for PanosterK. We found this to be not beneficial for PanosterS.

\begin{figure}[t]
\centering
  \includegraphics[width=0.48\textwidth]{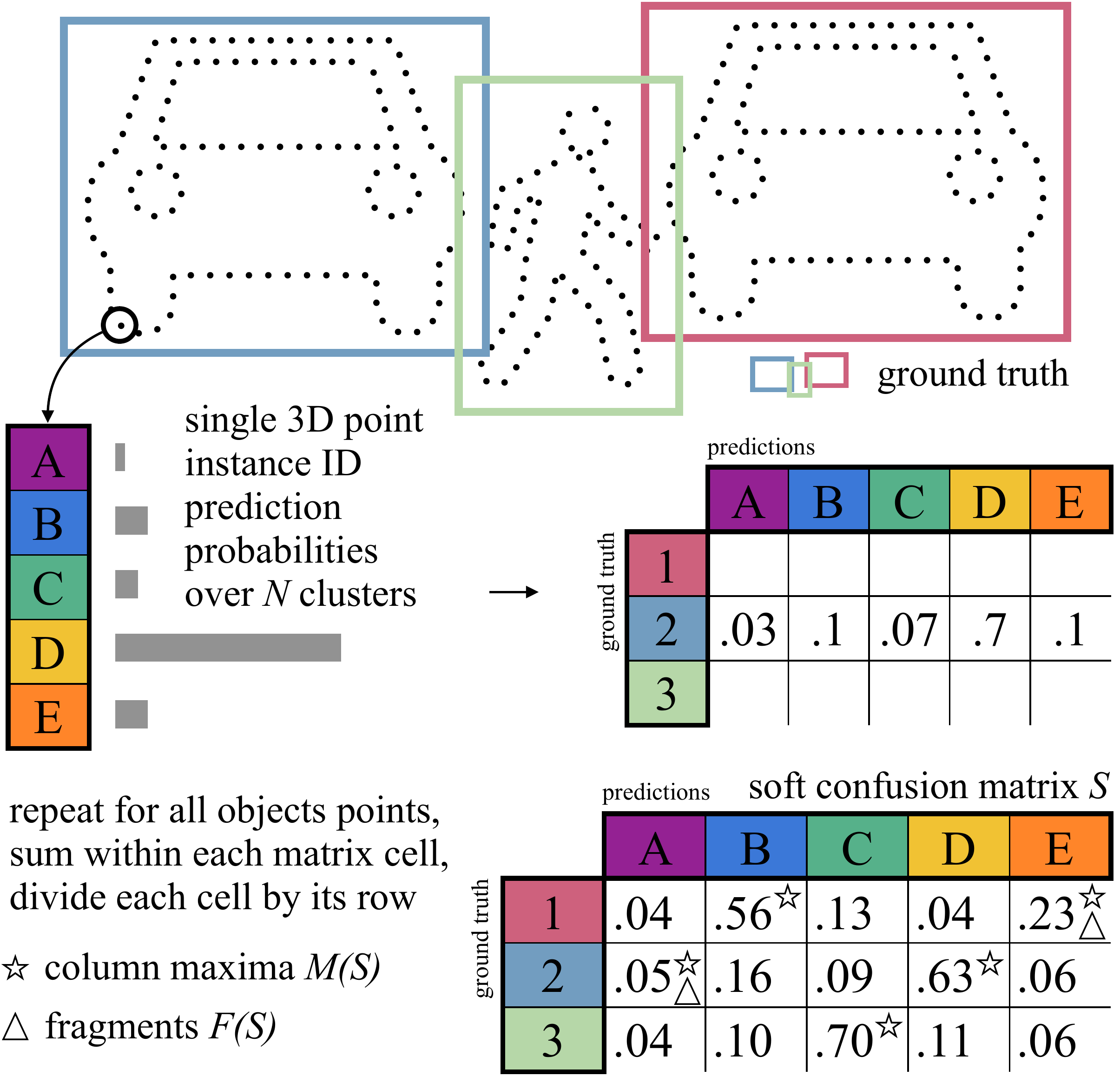}
   \caption{Construction of the matrix used for the computation of the instance clustering loss functions. The matrix $S$ is based on the instance branch segmentation probabilities, which are the cluster assignment probabilities. The figure shows a simplified scene with 3 instances and a model with $N=5$. Fragments $F(S)$ and all non-column maxima cells $Q(S)$ are minimized by fragmentation and impurity losses respectively.}
   \label{fig:matrix}
\end{figure}

\subsection{Learning-based Instance Clustering}
\label{sec:inst_seg}
Our instance branch predicts an instance ID for each point, including those belonging to \textit{stuff} regions, for which such ID has no meaning and will be later filtered out (Section~\ref{sec:pano_seg}). This instance ID is the output of the network after an \textit{argmax} operation over the instance branch \textit{logits}. We address instance segmentation as a class-agnostic segmentation task, with the underlying classification based on instance IDs, instead of the usual semantic classes. However, these IDs are interchangeable, similarly to clusters labels, since the essence of instance segmentation is grouping inputs to resemble object instances, e.g. instance clustering.
In this Section we describe how to predict such IDs.

Towards this end, we adapt a method originally proposed for separating pulses of radar signals by source, for aircraft applications~\cite{gasperini2020signal}.
In our LiDAR panoptic task, cluster elements are 3D points, and each ground truth cluster is an object instance, e.g. a car.
The idea is achieving instance segmentation by guiding the training optimization to minimize two complementary loss functions, which are extracted from a confusion matrix.
However, since confusion matrices are inherently non-differentiable, being based on \textit{argmax} operations over the class \textit{logits}, we construct suitable matrices by considering the probability distribution over the \textit{logits softmax} instead.
In particular, as depicted in Fig.~\ref{fig:matrix}, we build this soft confusion matrix $S$ by collecting the instance branch \textit{softmax} probabilities over the $N$ predictable instances, for all 3D points belonging to ground truth objects, and summing these probabilities together as follows.

For each LiDAR sweep, the \textit{softmax}-based confusion matrix $S$ is sized $G \times N$, with $G$ being the number of ground truth objects. As shown in Fig.~\ref{fig:matrix}, the matrix $S$ gathers the instance predictions made by the network over the input 3D points. The notation in this Section refers to sets of cells, rows and columns of the matrix $S$, rather than single 3D input points. The row $S_i$ corresponds to the $i$-th object, while the column $S^j$ represents the $j$-th predicted cluster. The cell $S_i^j$ value is the sum over the probability of each point of the $i$-th object to belong to the $j$-th predicted instance, divided by the sum of the row $S_i$ (i.e. the amount of 3D points constituting object $i$). Unlike in~\cite{gasperini2020signal}, where the matrix values are unbounded and depend on the clusters dimensions, we introduce this division transforming the matrix in percentage-based, to render Panoster robust against the imbalance of instance sizes.
The construction of $S$ is shown in Fig.~\ref{fig:matrix}.

\begin{figure}[t]
\centering
  \includegraphics[width=0.39\textwidth]{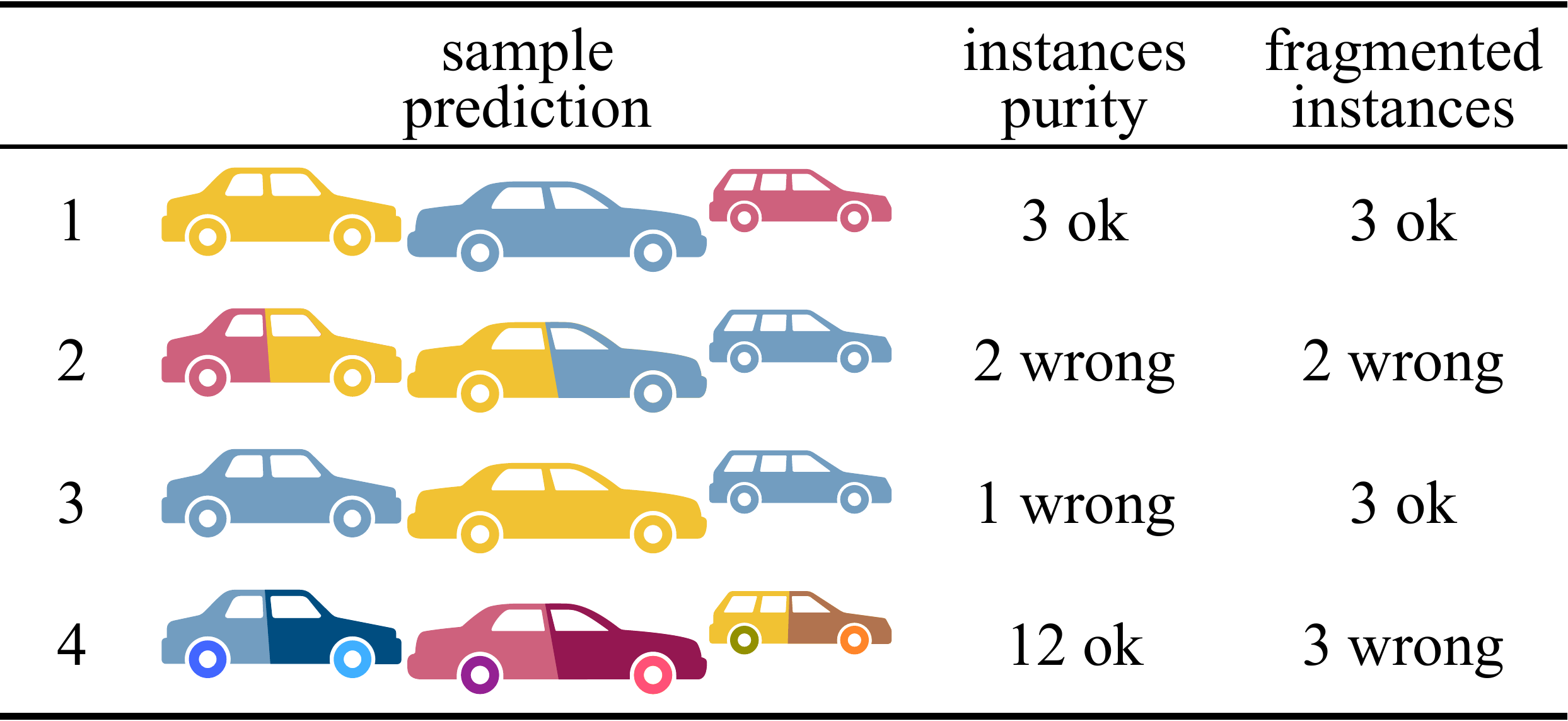}
   \caption{Sample instance segmentation predictions.
   In each row, colors represent predicted clusters for 3 cars. Apart for 1, 2-4 show sub-optimal results: in 2 objects are mixed and split in parts; in 3 the number of cars was underestimated, with impurity on the blue cluster, which merges two instances; in 4 the number of cars was overestimated, with severe fragmentation, as each object is split in four parts.}
   \label{fig:cars}
\end{figure}

This matrix $S$ is fully differentiable and can drive the network optimization. We deploy two loss functions computed from the matrix $S$, namely impurity and fragmentation. As shown in Fig.~\ref{fig:cars}, by aiming at both pure and non-fragmented instances, we can achieve the best outcome.

%

The \textbf{impurity loss} $L_{imp}$ in Eq.~\ref{eq:cil} tries to shift from 2 and 3 to 1 and 4 in Fig.~\ref{fig:cars}, by optimizing instances purity:

\begin{equation}
\begin{matrix}
    L_{imp} = \frac{\sum Q(S)}{\sum S}\\
    \scriptstyle Q(S):= \scriptstyle \left\{ S_i^j \right\} : \: S_i^j  \notin M^j(S) \:\: \forall i,j
    \label{eq:cil}
\end{matrix}
\end{equation}
where $M(S) := \left\{ \max \left( S ^ { j } \right) \forall~j \right\}$ is the set of column maxima of $S$, and $Q(S)$ is the set of all $S$ cells that are not part of $M(S)$. Column maxima $M(S)$ define each predicted instance by linking it with a ground truth object. By minimizing $Q(S)$, this loss increases the correspondence between ideal and predicted instances.

The additional \textbf{fragmentation loss} $L_{fr}$ 
reduces fragmentation. It limits overestimating the amount of objects, by penalizing the $S$ cells responsible for fragments $F(S)$, which occur when two or more column maxima $M(S)$ fall within the same row:
\begin{equation}
\begin{matrix}
    L_{fr} = \frac{\sum F(S) \, \oslash \, F(S)}{N}\\
    \scriptstyle F(S) \, := \, \scriptstyle \left\{ S_i^j \right\} : \: S_i^j  \in M_i(S) \; \land \; S_i^j \, \neq \, \scriptstyle \max \left( M_i (S) \right) \:\: \forall i,j \\
    \scriptstyle M_i(S) \, := \, \scriptstyle S_i \; \cap \; M(S)
    \label{eq:cfl}
\end{matrix}
\end{equation}
In the equation, the numerator is the differentiable equivalent of counting the fragments $F(S)$, thanks to the Hadamard element-wise division $\oslash$.
Therefore, this loss enforces that each ground truth object corresponds to only one predicted instance, bringing use case 4 of Fig.~\ref{fig:cars} to 1.

Our instance branch is trained with a combination of $L_{imp}$ and $L_{fr}$, weighted 0.2 and 0.05 respectively.

\begin{table*}[t]
    \begin{center}
    \caption{Panoptic segmentation comparison on test set results of SemanticKITTI. All values in [\%] with \textit{St} for \textit{stuff} and \textit{Th} for \textit{thing}.}
        \begin{tabular}
            {l|cccc|ccc|ccc|c}
            \toprule
            Method & PQ & PQ$^\dagger$ & SQ & RQ & PQ\textsuperscript{Th} & SQ\textsuperscript{Th} & RQ\textsuperscript{Th} & PQ\textsuperscript{St} & SQ\textsuperscript{St} & RQ\textsuperscript{St} & mIoU\\ 
            \midrule
            
            RangeNet++~\cite{milioto2019rangenet++} + PointPillars~\cite{lang2019pointpillars} & 37.1 & 45.9 & 75.9 & 47.0 & 20.2 & 75.2 & 25.2 & 49.3 & 76.5 & 62.8 & 52.4 \\
            
            LPSAD~\cite{milioto2020lidar} & 38.0 & 47.0 & 76.5 & 48.2 & 25.6 & 76.8 & 31.8 & 47.1 & 76.2 & 60.1 & 50.9 \\
            
            KPConv~\cite{thomas2019kpconv} + PointPillars~\cite{lang2019pointpillars} & 44.5 & 52.5 & 80.0 & 54.4 & 32.7 & 81.5 & 38.7 & 53.1 & \textbf{79.0} & 65.9 & 58.8 \\
            \midrule
            
            PanosterK~[Ours] & 45.6 & 52.8 & 78.1 & 57.0 & 32.4 & 77.1 & 41.6 & \textbf{55.1} & 78.8 & \textbf{68.2} & \textbf{60.4}\\
            
            PanosterK + \textit{post\_*}~[Ours] & \textbf{52.7} & \textbf{59.9} & \textbf{80.7} & \textbf{64.1} & \textbf{49.4} & \textbf{83.3} & \textbf{58.5} & \textbf{55.1} & 78.8 & \textbf{68.2} & 59.9\\
            \bottomrule
        \end{tabular}
\label{tab:test_res}
    \end{center}
\vspace{-0.25cm}
\end{table*}

\begin{table*}[t]
    \setlength\tabcolsep{3.7pt}
    \caption{Detailed class-wise PQ results of SemanticKITTI test set. All values in [\%].}
    \begin{center}
        \begin{tabular}
            {lcccccccccccccccccccc}
            \toprule
            Method & PQ & \begin{sideways}car\end{sideways} & \begin{sideways}truck\end{sideways} & \begin{sideways}bicycle\end{sideways} & \begin{sideways}motorcycle\end{sideways} & \begin{sideways}other vehicle\end{sideways} & \begin{sideways}person\end{sideways} & \begin{sideways}bicyclist\end{sideways} & \begin{sideways}motorcyclist\end{sideways} & \begin{sideways}road\end{sideways} & \begin{sideways}sidewalk\end{sideways} & \begin{sideways}parking\end{sideways} & \begin{sideways}other ground\end{sideways} & \begin{sideways}building\end{sideways} & \begin{sideways}vegetation\end{sideways} & \begin{sideways}trunk\end{sideways} & \begin{sideways}terrain\end{sideways} & \begin{sideways}fence\end{sideways} & \begin{sideways}pole\end{sideways} & \begin{sideways}traffic sign\end{sideways} \\ 
            \midrule
            RangeNet++ + PointP. & 37.1 & 66.9 & 6.7 & 3.1 & 16.2 & 8.8 & 14.6 & 31.8 & 13.5 & \textbf{90.6} & \textbf{63.2} & \textbf{41.3} & 6.7 & 79.2 & 71.2 & 34.6 & 37.4 & 38.2 & 32.8 & 47.4\\
            KPConv + PointP. & 44.5 & 72.5 & 17.2 & 9.2 & 30.8 & 19.6 & 29.9 & 59.4 & 22.8 & 84.6 & 60.1 & 34.1 & \textbf{8.8} & 80.7 & 77.6 & 53.9 & \textbf{42.2} & \textbf{49.0} & 46.2 & 46.8\\
            \midrule
            PanosterK & 45.6 & 49.6 & 13.7 & 9.2 & 33.8 & 19.5 & 43.0 & 50.2 & 40.2 & 90.2 & 62.5 & 34.5 & 6.1 & \textbf{82.0} & \textbf{77.7} & \textbf{55.7} & 41.2 & 48.0 & \textbf{48.9} & \textbf{59.8}\\
            PanosterK + \textit{post\_*} & \textbf{52.7} & \textbf{84.0} & \textbf{18.5} & \textbf{36.4} & \textbf{44.7} & \textbf{30.1} & \textbf{61.1} & \textbf{69.2} & \textbf{51.1} & 90.2 & 62.5 & 34.5 & 6.1 & \textbf{82.0} & \textbf{77.7} & \textbf{55.7} & 41.2 & 48.0 & \textbf{48.9} & \textbf{59.8}\\
            \bottomrule
        \end{tabular}
\label{tab:test_pq}
    \end{center}
\end{table*}

\subsection{Panoptic Segmentation}
\label{sec:pano_seg}
PS requires a semantic class for each point, as well as instance IDs for \textit{thing} objects.
Both branches output a prediction for each point of \textit{stuff} and \textit{thing} classes. 
However, instance IDs are meaningless for \textit{stuff} points, and will be filtered out as follows.
As shown in Fig.~\ref{fig:framework}, we extract an object mask from the semantic prediction, distinguishing \textit{things} from \textit{stuff} points. We use this mask to discard instance predictions for \textit{stuff}, and retain instance IDs for \textit{things}. 
Finally, PS predictions are obtained by stacking semantic classes and instance IDs.

\textbf{Optional post-processing.} As described in Section~\ref{sec:inst_seg}, unlike previous works~\cite{cheng2020panoptic, milioto2020lidar}, our instance predictions do not necessarily require further grouping or post-processing.
Nevertheless, we explored three different heuristic strategies, as plausibility checks, exploiting 3D information and the synergy between the two tasks.
To improve detection and purity, DBSCAN~\cite{ester1996dbscan} targets the third use case of Fig.~\ref{fig:cars}. We call this strategy \textit{post\_splitter}: it splits instances which are too large in space w.r.t. their corresponding predicted semantic classes.
Furthermore, to reduce fragmentation, \textit{post\_merger} can be deployed to fix the fourth use case of Fig.~\ref{fig:cars}. It uses DBSCAN to cluster the 3D centers of those instances of the same semantic class, which are too close to each other.
Finally, with \textit{post\_cyclists}, we turn \textit{bicyclists} who have no \textit{bicycle} nearby, but a \textit{motorbike}, into \textit{motorcyclists}.



\section{Experiments}
\label{sec:experiments}

\subsection{Experimental Setup}
\label{sec:setup}

\textbf{Dataset.} We evaluate on the challenging Semantic-KITTI~\cite{behley2019skitti}, which was recently extended~\cite{behley2020benchmark} by providing instance IDs for all LiDAR scans of KITTI odometry~\cite{geiger2012kitti}. It contains 23201 full 360 degree scans for training and 20351 for testing, sampled from 22 sequences from various German cities. We use training sequence 08 as validation split, following the convention. Annotations comprise point-wise labels within 50 meters radius across 22 classes, 19 of which evaluated on the test server: 11 \textit{stuff} and 8 \textit{thing}.

\textbf{Model training.}
Experiments are focused mainly on PanosterK, based on KPConv~\cite{thomas2019kpconv}, while showing the flexibility of Panoster by applying it also on SalsaNext~\cite{cortinhal2020salsanext} with PanosterS.
Unless otherwise noted, we used the same training parameters as in~\cite{thomas2019kpconv} for PanosterK and~\cite{cortinhal2020salsanext} for PanosterS, with the addition of our instance clustering loss functions.
To better fit the PS task, we used input point clouds covering the entire scene, instead of small regions as in~\cite{thomas2019kpconv}.
At each training iteration, we fed PanosterK with at most 80K points.
PanosterK used rigid KPConv kernels. For PanosterS we used batch size 4 and learning rate 0.001.
We implemented our method in PyTorch and trained the models on a single NVIDIA Tesla V100 32GB GPU. We trained PanosterK from scratch for at least 200K iterations. PanosterS was pretrained on semantic segmentation, and fine-tuned on panoptic segmentation for 100K iterations. The PanosterK model submitted to the test server was trained on the entire training set, including the validation set.

\textbf{Inference.}
Unlike KPConv~\cite{thomas2019kpconv}, at inference time we fed the whole 50 meters radius scene in a single forward pass, without any test-time augmentation.

\textbf{Evaluation metrics.}
We report mean IoU (mIoU) and panoptic quality (PQ)~\cite{kirillov2019panoptic} to evaluate semantic and panoptic segmentation, both averaged over all classes. PQ can be seen as the multiplication of segmentation quality (SQ) and recognition quality (RQ)~\cite{kirillov2019panoptic}, with the former being the mIoU of matched segments and the latter representing the $F_1$ score, commonly adopted for object detection. Additionally, we report the alternative PQ$^\dagger$~\cite{porzi2019seamless}, which ignores RQ for \textit{stuff} classes.
Since our main focus and contribution within PS is on instance segmentation, among the various metrics we concentrate on RQ\textsuperscript{Th}, i.e. the $F_1$ score for \textit{thing} classes, which is more appropriate than PQ to compare instance segmentation performances.

\subsection{Panoptic Segmentation}
We compared Panoster with other published PS approaches for LiDAR point clouds, and submitted PanosterK to the SemanticKITTI challenge.
Two strong baselines were proposed in~\cite{behley2020benchmark} combining KPConv~\cite{thomas2019kpconv} or RangeNet++~\cite{milioto2019rangenet++} with PointPillars~\cite{lang2019pointpillars}, semantic segmentation and object detection methods. Furthermore, we compared with PanopticTrackNet~\cite{hurtado2020mopt} and LPSAD~\cite{milioto2020lidar}, top-down and bottom-up approaches respectively.

Table~\ref{tab:test_res} shows results on the test set.
Overall, PanosterK delivered superior performance compared to previously published approaches.
Despite obtaining worse semantics with a lower SQ, and significantly lower SQ\textsuperscript{Th} which penalized PQ\textsuperscript{Th}, PanosterK's strengths are its finer instance segmentation abilities, shown on \textit{thing} classes, resulting in a notable 2.9 increase in RQ\textsuperscript{Th} on top of state-of-the-art KPConv and PointPillars, albeit sharing the same backbone.
Considering individual classes, Table~\ref{tab:test_pq} shows that PointPillars~\cite{lang2019pointpillars}, which is a strong object detector, performed better than our instance clustering approach on the classes \textit{car}, \textit{truck} and \textit{bicyclist}, while ours prevailed on \textit{person}, \textit{motorcycle} and \textit{motorcyclist}. Nevertheless, compared to the combined approaches, our PanosterK was able to deliver a better panoptic segmentation, while using a single network in a multi-task fashion.
Interestingly, as shown in Table~\ref{tab:test_pq}, although PanosterK performed comparably to previous approaches on most amorphous \textit{stuff} classes in terms of PQ, it outperformed them on all those \textit{stuff} classes which could be split into instances, such as \textit{traffic sign}, \textit{pole}, and \textit{trunk}. This could be due to our instance losses, which enforced learning the concept of object, despite not being applied on \textit{stuff} regions.
Furthermore, LPSAD~\cite{milioto2020lidar}, the only other bottom-up method to date,
despite a more complex architecture
and a necessary grouping step,
achieved significantly lower PQ and RQ\textsuperscript{Th}.
Overall, we attribute PanosterK performance to the high quality instance segmentation resulting from its built-in learning-based separation approach. By optimizing the cluster assignments directly for purity and non-fragmentation, PanosterK could achieve good instance segmentation, without affecting semantic accuracy.
Furthermore, applying our optional post-processing techniques \textit{post\_*} (Section~\ref{sec:pano_seg}) increased PQ by 7.1 reaching 52.7, with RQ\textsuperscript{Th} at 58.5, improving all \textit{thing} classes.
More details on this are presented in Section~\ref{sec:ablative}.

We report validation set results in Table~\ref{tab:val_res}, including PanopticTrackNet~\cite{hurtado2020mopt}, which has no test entry to date. We add validation results of the combined approaches~\cite{behley2020benchmark}, as outlined in~\cite{hurtado2020mopt}. Although our projection-based PanosterS, outperformed previous methods, our point-based PanosterK delivered even better results. We attribute this to the direct consumption of 3D points, which leads to a better use of geometric information, fundamental when separating instances in 3D space.


\begin{table}
    \begin{center}
    \caption{Panoptic segmentation on the validation set of SemanticKITTI. All values in [\%].}
        \begin{tabular}
            {l|ccc|cc|c}
            \toprule
            Method & PQ & SQ & RQ & PQ\textsuperscript{Th} & RQ\textsuperscript{Th} & mIoU\\
            \midrule
            RangeNet++ + Pt.P. & 36.5 & 73.0 & 44.9 & 19.6 & 24.9 & 52.8\\
            
            LPSAD & 36.5 & - & - & - & 28.2 & 50.7\\
            
            PanopticTrackNet & 40.0 & 73.0 & 48.3 & 29.9 & 33.6 & 53.8\\
            
            KPConv + Pt.P. & 41.1 & 74.3 & 50.3 & 28.9 & 33.1 & 56.6\\
            
            \midrule
            
            PanosterS & 43.5 & 68.6 & 55.2 & 32.6 & 42.3 & 57.8\\
            
            PanosterS + \textit{post\_*} & 51.1 & 76.9 & 62.1 & 50.5 & 58.6 & 57.8\\
            
            PanosterK & 48.4 & 73.0 & 60.1 & 39.5 & 50.0 & 60.4\\
            
            PanosterK + \textit{post\_*} & \textbf{55.6} & \textbf{79.9} & \textbf{66.8} & \textbf{56.6} & \textbf{65.8} & \textbf{61.1}\\
            \bottomrule
        \end{tabular}
\label{tab:val_res}
    \end{center}
\end{table}

\begin{table}
    \begin{center}
    \caption{Ablation study on the validation set of SemanticKITTI, with \textit{c} for \textit{car}, \textit{b} for \textit{bicycle} and \textit{p} for \textit{person}. All values in [\%].}
        \begin{tabular}
            {l|cc|ccc|c}
            \toprule
            Method & PQ & RQ\textsuperscript{Th} & RQ\textsuperscript{c} & RQ\textsuperscript{b} & RQ\textsuperscript{p} & mIoU\\
            \midrule
            
            
            KPConv + instance & 45.9 & 45.2 & 53.3 & 42.4 & 51.2 & 58.3\\
            + extra decoder & 45.2 & 43.5 & 65.4 & 37.5 & 60.8 & 59.0\\
            + \% conf. matrix & 47.6 & 49.3 & 75.5 & 47.1 & 67.1 & 60.0\\
            + Lov\`{a}sz-softmax & 48.3 & 49.6 & 74.2 & 38.9 & 70.1 & \textbf{61.1}\\
            + small inst. 3x & 48.5 & 49.4 & 75.0 & 40.0 & 75.1& 60.2\\
            + skip=\textbf{PanosterK} & 48.4 & 50.0 & 79.5 & 50.6 & 74.1 & 60.4\\
            \midrule
            + \textit{post\_splitter} & 50.4 & 54.4 & 87.4 & 60.1 & 85.4 & 60.4\\
            + \textit{post\_merger} & 54.7 & 62.9 & \textbf{95.3} & \textbf{60.9} & \textbf{87.6} & 60.4\\
            + \textit{post\_cyclists} & \textbf{55.6} & \textbf{65.8} & \textbf{95.3} & \textbf{60.9} & \textbf{87.6} & \textbf{61.1}\\
            \bottomrule
        \end{tabular}
\label{tab:ablative}
    \end{center}
\end{table}

\subsection{Ablation Study}
\label{sec:ablative}

In Table~\ref{tab:ablative}, we summarize an ablation study to assess the impact of different components of PanosterK on PS.

\textbf{Instance improvements}.
Although using a single decoder achieved higher PQ and RQ\textsuperscript{Th}, dedicated decoders led to a significantly better RQ on \textit{car} and \textit{person}, while improving semantic performance on mIoU, motivating our choice for the latter configuration.
As described in Section~\ref{sec:inst_seg}, compared to the instance clustering approach adapted from~\cite{gasperini2020signal}, we increased the robustness against unbalanced cluster sizes.
This change, shown in Table~\ref{tab:ablative} as "\% conf. matrix", largely improved the predictions, preventing closer and larger objects, in terms of point counts, from dominating smaller and further ones in the clustering process.
Furthermore, we added a skip connection from semantic to instance head, which led to a higher RQ\textsuperscript{Th}, with better detection for the classes \textit{car} and \textit{bicycle}.

\begin{figure}[t]
\centering
  \includegraphics[width=0.48\textwidth]{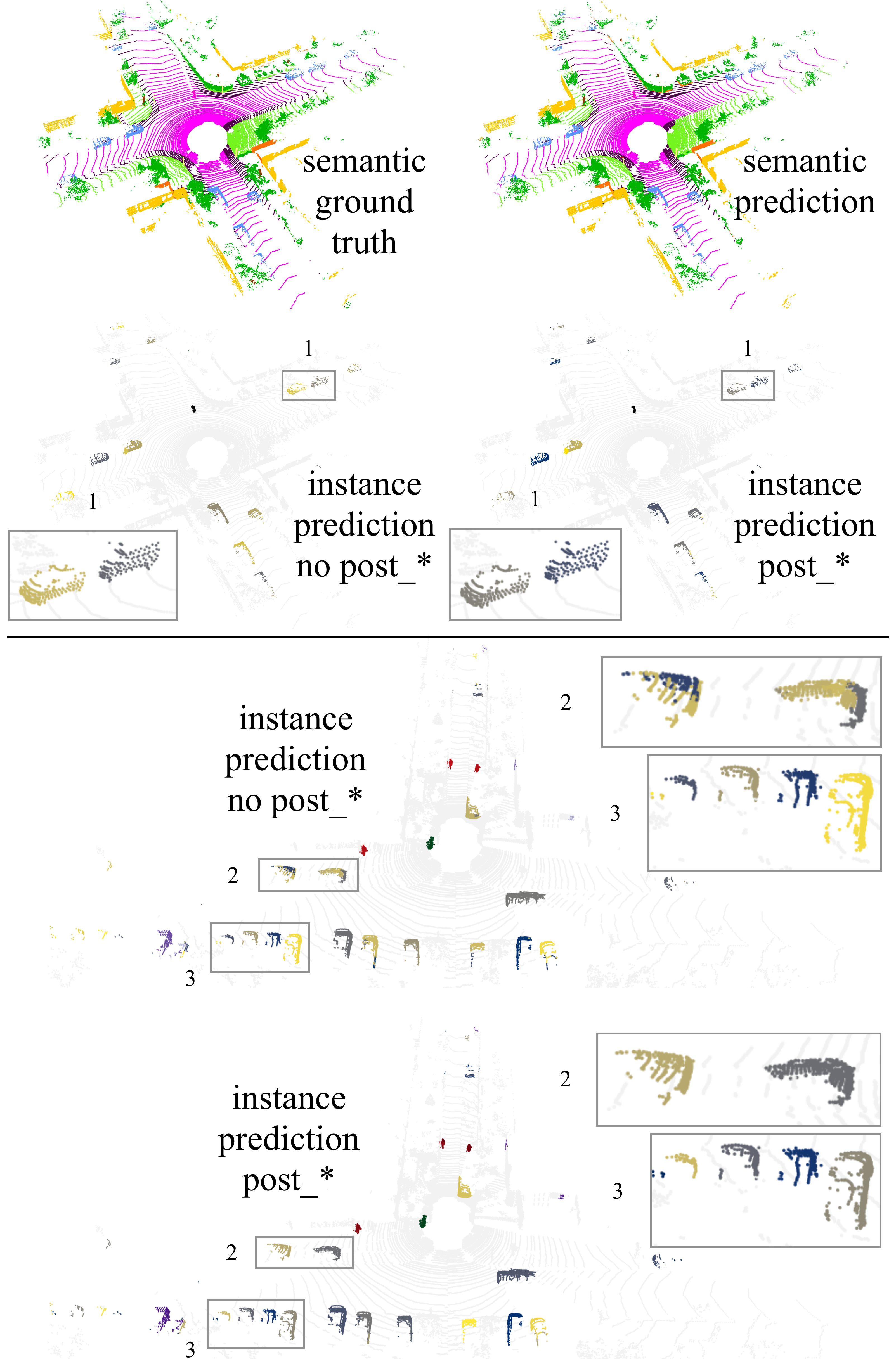}
   \caption{Qualitative results of PanosterK on two challenging samples from the validation set of SemanticKITTI. Enlarged numbered boxes show representative regions.}
   \label{fig:qualitative}
\end{figure}

\begin{figure*}[th!]
\centering
  \includegraphics[width=1.\textwidth]{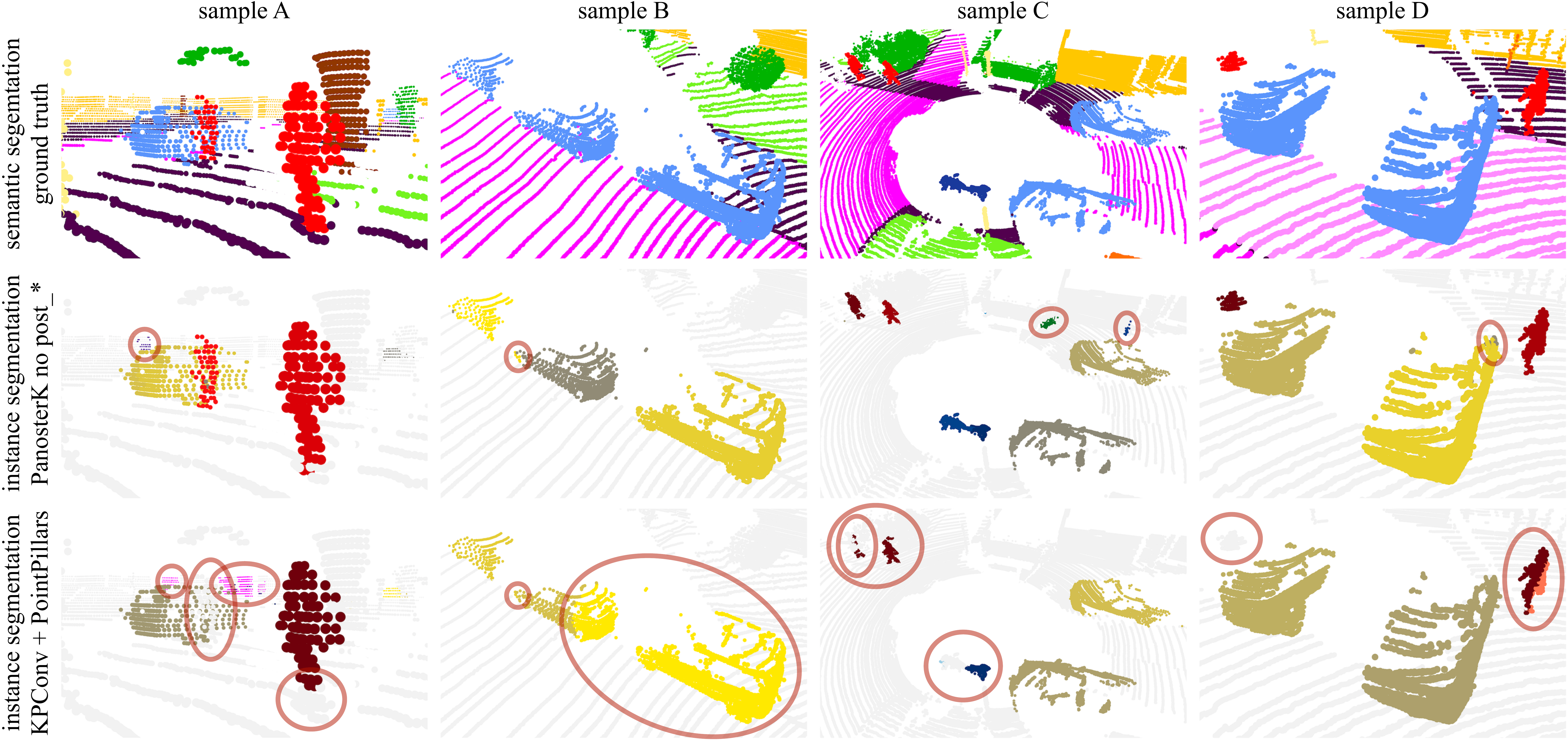}
   \caption{Qualitative comparison on four challenging portions of the validation set of SemanticKITTI, with instance segmentation predicted by our PanosterK without post-processing, and KPConv + PointPillars. In the second and third rows, each color shade represents an ID, while red circles indicate errors.}
   \label{fig:qualitative_comparison}
\end{figure*}

\textbf{Semantic improvements}. Since Panoster is a bottom-up method, to avoid errors propagating through the \textit{things} mask, precise semantic segmentation is key for high quality panoptic outputs. Towards this end, we applied the Lov\`{a}sz-softmax loss and we limited false negatives by increasing the weight of small instances (denoted by "small inst. 3x" in Table~\ref{tab:ablative}). The former had a positive impact on mIoU, PQ and RQ\textsuperscript{Th}. The latter, despite decreasing mIoU, increased precision on smaller \textit{thing} classes, such as \textit{person}.

\textbf{Number of predictable clusters N}. For all experiments we set N equal to 60, higher than the amount of objects found across the dataset in a single scan. This number can be seen equivalent to the amount of bounding boxes processed in an object detector~\cite{he2017mask}. 
Two architectures using N = \{60, 90\} resulted in a PQ difference of 0.1,
showing Panoster insensitivity against N.

\textbf{Post-processing}. As it can be seen in Tables~\ref{tab:test_res},~\ref{tab:test_pq},~\ref{tab:val_res} and~\ref{tab:ablative}, our optional \textit{post\_*} strategies (Section~\ref{sec:pano_seg}) bring significant improvements to Panoster predictions.
They complement our model, fixing its clustering errors through plausibility checks by exploiting 3D data and the task duality of PS. The largest improvement is brought by \textit{post\_merger}, which joins close instances likely to be part of the same object, divided by the model or by \textit{post\_splitter}.

\subsection{Runtime Comparison}

Our PanosterK model runs on 80K points at 58 FPS at inference time for a single forward-pass on a NVIDIA GTX 1080 8GB GPU, which is 3 FPS (i.e. 5\%) slower than its semantic counterpart. Therefore, Panoster brings panoptic capabilities with a relatively small overhead with respect to semantic segmentation networks.
This high speed is possible thanks to the simplifications we adopted at inference time (Section~\ref{sec:setup}).
In comparison, LPSAD, the previously fastest existing method for PS on LiDAR point clouds, runs at 11.8 FPS~\cite{milioto2020lidar}. Panoster's built-in learning-based clustering avoids cumbersome grouping strategies to form instances, allowing it to be fast. Moreover, Panoster is more efficient than the combined methods, which require three networks, i.e., one for semantic segmentation, and two for detecting smaller and larger objects~\cite{behley2020benchmark}.

\subsection{Qualitative Results}
\label{sec:qualitative}

In Fig.~\ref{fig:qualitative} we showcase PanosterK predictions on two complex validation scenes, with road intersections and several objects. Each shows the whole 50m radius scan. Despite point sparsity, predictions do not degenerate by increasing the distance from the sensor. Additionally, Panoster is able to distinguish neighboring objects, assigning them different IDs (i.e. colors), as in detail 3 of Fig.~\ref{fig:qualitative}. Although in challenging scenes, such as the bottom one in the figure, it reuses the same ID for multiple instances, these can be fixed via post-processing as matching IDs are found only in distant objects. We attribute this behavior to the relatively low amount of instances and complex scenes in most training samples.

Furthermore, in Fig.~\ref{fig:qualitative_comparison} we compare PanosterK, without any post-processing, against the approach proposed in~\cite{behley2020benchmark} combining KPConv and PointPillars. Qualitative results confirm the findings of our experiments, with our PanosterK delivering superior instance segmentation, with minor errors. Despite relying on a strong object detector, the combined approach completely ignored the challenging pedestrian in sample A next to the car and the partially occluded one in D, wrongly divided a person in D, partially ignored the left person and the rare motorcycle in C, assigned the same ID to multiple people in C, and wrongly detected parked cars in B.
As both methods are based on the same semantic architecture, i.e. KPConv, we attribute this difference to the benefit of the integrated panoptic solution offered by our Panoster, as well as to the issues rising from dealing with multiple and possibly contrasting predictions when combining the outputs of various networks.



\section{Conclusion}
\label{sec:concl}

In this paper we proposed Panoster: a fast, flexible, and proposal-free panoptic segmentation method, which we evaluated on the challenging LiDAR-based SemanticKITTI dataset, across two different input representations.
Panoster outperformed published state-of-the-art approaches, while being simpler and faster. Our method directly delivers instance IDs with a learning-based clustering solution, embedded in the model and optimized for pure and non-fragmented clusters.
With its small overhead, Panoster constitutes a valid end efficient approach to extend existing and upcoming semantic methods to perform panoptic segmentation.


\bibliographystyle{IEEEbib}
\bibliography{refs}

\end{document}